\documentclass{article}

\usepackage{microtype}
\usepackage{graphicx}
\usepackage{subfigure}
\usepackage{booktabs} 

\usepackage{hyperref}

\usepackage[accepted]{icml2024}

\usepackage{amsmath}
\usepackage{amssymb}
\usepackage{mathtools}
\usepackage{amsthm}
\usepackage{bm}

\usepackage[capitalize,noabbrev]{cleveref}

\theoremstyle{plain}

\theoremstyle{definition}

\theoremstyle{remark}

\usepackage[textsize=tiny]{todonotes}

\icmltitlerunning{Continual Deep Learning on the Edge via Stochastic Local Competition
among Subnetworks}

\begin{document}

\twocolumn[
\icmltitle{Continual Deep Learning on the Edge via Stochastic Local Competition
among Subnetworks}



\icmlsetsymbol{equal}{*}

\begin{icmlauthorlist}
\icmlauthor{Theodoros Christophides}{cut}
\icmlauthor{Kyriakos Tolias}{ekt}
\icmlauthor{Sotirios Chatzis}{cut}
\end{icmlauthorlist}

\icmlaffiliation{cut}{Department of Electrical Eng., Computer Eng., and Informatics; Cyprus University of Technology}
\icmlaffiliation{ekt}{National Documentation Center, Greece}

\icmlcorrespondingauthor{Sotirios Chatzis}{sotirios.chatzis@cut.ac.cy}

\icmlkeywords{Machine Learning, ICML}

\vskip 0.3in
]



\printAffiliationsAndNotice{\icmlEqualContribution} 

\begin{abstract}
Continual learning on edge devices poses unique challenges due to
stringent resource constraints. This paper introduces a novel method
that leverages stochastic competition principles to promote sparsity,
significantly reducing deep network memory footprint and computational
demand. Specifically, we propose deep networks that comprise blocks
of units that compete locally to win the representation of each arising
new task; competition takes place in a stochastic manner. This type
of network organization results in sparse task-specific representations
from each network layer; the sparsity pattern is obtained during training
and is different among tasks. Crucially, our method sparsifies both
the weights and the weight gradients, thus facilitating training on
edge devices. This is performed on the grounds of winning probability
for each unit in a block. During inference, the network retains only
the winning unit and zeroes-out all weights pertaining to non-winning
units for the task at hand. Thus, our approach is specifically tailored
for deployment on edge devices, providing an efficient and scalable
solution for continual learning in resource-limited environments. 
\end{abstract}

\section{Introduction}

\label{intro} Continual Learning (CL), also referred to as Lifelong
Learning \cite{Thrun2}, aims to learn sequential tasks and acquire
new information while preserving knowledge from previous learned tasks
\cite{Thrun}. This paper's focus is on a variant of CL dubbed class-incremental
learning (CIL) \cite{Belouadah,Gupta,Deng}. The main principle of
CIL is a CL scenario where on each iteration we are dealing with data
from a specific task, and each task contains new classes that must
be learnt.

Edge devices, characterized by their limited computational resources,
necessitate efficient machine learning models to perform tasks effectively.
Sparsity in neural networks emerges as a critical feature to address
these limitations, reducing memory requirements and computational
costs. This work introduces a stochastic competition mechanism to
induce sparsity, optimizing continual learning processes specifically
for such constrained environments.

Recently, different research groups have drawn inspiration from the
lottery ticket hypothesis (LTH) \cite{Frankle} to introduce the lifelong
tickets (LLT) method \cite{Chen_Zhang}, the Winning SubNetworks (WSN)
method \cite{Kang}, and, more recently, the Soft-SubNetworks approach
\cite{Kang_Yoon}. However, these recent advances are confronted with
major limitations: (i) LLT entails an iterative pruning procedure,
that requires multiple repetitions of the training algorithm for each
task; this is not suitable for edge devices. (ii) The existing alternatives
do not take into consideration the uncertainty in the used datasets,
which would benefit from the subnetwork selection process being stochastic,
as opposed to hard unit pruning. In fact, it has been recently shown
that stochastic competition mechanisms among locally competing units
can offer important generalization capacity benefits for deep networks
used in as diverse challenges as adversarial robustness \cite{Panousis_Chatzis_Adv},
video-to-text translation \cite{Voskou_Chatzis}, and model-agnostic
meta-learning \cite{Kalais_Chatzis}.

In a different vein, SparCL \cite{Wang:2022aa} has been the first
work on CL specifically designed to tackle applications on edge devices,
where resource constraints are significant. Apart from weight sparsity,
SparCL also considers data efficiency and gradient sparsity to accelerate
training while preserving accuracy. SparCL dynamically maintains important
weights for current and previous tasks and adapts the sparsity during
transitions between tasks, which helps in mitigating catastrophic
forgetting---a common challenge in continual learning. Importantly,
in contrast to LTH-based methods, the foundational characteristics
of the approach introduce sparsity into the gradient updates, which
reduces the computational cost during backpropagation. This component
is critical for efficient training on hardware with limited computational
capabilities, such as mobile phones or other edge devices.

Inspired from these facts, this work proposes a radically different
regard toward addressing catastrophic forgetting in CIL. Our approach
is founded upon the framework of stochastic local competition which
is implemented in a task-wise manner. Specifically, our proposed approach
relies upon the following novel contributions: 
\begin{itemize}
\item \textbf{Task-specific sparsity in the learned representations.} We
propose a novel mechanism that inherently learns to extract sparse
task-specific data representations. Specifically, each layer of the
network is split into blocks of competing units; local competition
is stochastic and it replaces traditional nonlinearities, e.g. ReLU.
Being presented with a new task, each block learns a distribution
over its units that governs which unit specializes in the presented
task. We dub this type of nonlinear units as \textit{task winner-takes-all
(TWTA)}. Under this scheme, the network learns a Categorical posterior
over the competing block units; this is the winning unit posterior
of the block. Only the winning unit of a block generates a non-zero
output fed to the next network layer. This renders sparse the generated
representations, with the sparsity pattern being task-specific. 
\item \textbf{Weight gradient pruning driven from the learned stochastic
competition posteriors.} During training, the network utilizes the
learned Categorical posteriors over winning block units to introduce
sparsity into the gradient updates. In a sense, the algorithm inherently
masks out the gradient updates of the block units with lower winning
posteriors. This is immensely important when deep network training
is carried out on edge devices. 
\item \textbf{Winner-based weight pruning at inference time.} During inference
for a given task, we use the (Categorical) winner posteriors learned
for the task to select the winner unit of each block; we zero-out
the remainder block units. This forms a \textit{task-winning ticket}
used for inference. This way, the size of the network used at inference
time is significantly reduced; pruning depends on the number of competing
units per block, since we drop all block units except for the selected
winner with maximum winning posterior. 
\end{itemize}
We evaluate our approach, dubbed TWTA for CIL (\textit{TWTA-CIL}),
on image classification benchmarks. We show that our approach shows
superior performance compared to both conventional CL methods and
CL-adapted sparse training methods on all benchmark datasets. This
leads to a (i) considerable improvement in accuracy, while (ii) yielding
task-specific networks that require immensely less FLOPs and impose
a considerably lower memory footprint.

The remainder of this paper is organized as follows: In Section \ref{approach},
we introduce our approach and describe the related training and inference
processes. Section \ref{related_work} briefly reviews related work.
In Section \ref{experiments}, we perform an extensive experimental
evaluation and ablation study of the proposed approach. In the last
Section, we summarize the contribution of this work.

\section{Proposed Approach}

\label{approach}

\subsection{Problem Definition}

CIL objective is to learn a unified classifier from a sequential stream
of data comprising different tasks that introduce new classes. CIL
methods should scale to a large number of tasks without immense computational
and memory growth. Let us consider a CIL problem $T$ which consists
of a sequence of $n$ tasks, $T=\{(C^{(1)},D^{(1)}),(C^{(2)},D^{(2)}),\dots,(C^{(n)},D^{(n)})\}$.
Each task $t$ contains data $D^{(t)}=(\bm{x}^{(t)},\bm{y}^{(t)})$
and new classes $C^{(t)}=\{c_{m_{t-1}+1},c_{m_{t-1}+2},\dots,c_{m_{t}}\}$,
where $m_{t}$ is the number of presented classes up to task $t$.
We denote as $\bm{x}^{(t)}$ the input features, and as $\bm{y}^{(t)}$
the one-hot label vector corresponding to $\bm{x}^{(t)}$.

When training for the $t$-th task, we use the data of the task, $D^{(t)}$.
We consider learners-classifiers that are deep networks parameterized
by weights $\bm{W}$, and we use $f(\bm{x}^{(t)};\bm{W})$ to indicate
the output Softmax logits for a given input $\bm{x}^{(t)}$. Facing
a new dataset $D^{(t)}$, the model's goal is to learn new classes
and maintain performance over old classes.

\subsection{Model formulation}

\label{model_form}

Our approach integrates a stochastic competition mechanism within
the training process to promote sparsity. By selectively activating
a subset of neurons and pruning less important connections, our method
maintains a high level of accuracy while significantly reducing the
model's complexity. This sparsity not only ensures lower memory usage
but also accelerates inference, making it ideal for edge devices.

Let us denote as $\bm{x}^{(t)}\in\mathbb{R}^{E}$ an input representation
vector presented to a dense ReLU layer of a traditional deep neural
network, with corresponding weights matrix $\bm{W}\in\mathbb{R}^{E\times K}$.
The layer produces an output vector $\bm{y}^{(t)}\in\mathbb{R}^{K}$,
which is fed to the subsequent layers.

In our approach, a group of $J$ ReLU units is replaced by a group
of $J$ competing linear units, organized in one block; each layer
contains $I$ blocks of $J$ units. Within each block, different units
are specialized in different tasks; only one block unit specializes
in a given task $t$. The layer input is now presented to each block
through weights that are organized into a three-dimensional matrix
$\bm{W}\in\mathbb{R}^{E\times I\times J}$. Then, the $j$-th ($j=1,\dots,J$)
competing unit within the $i$-th ($i=1,\dots,I$) block computes
the sum $\sum_{e=1}^{E}(w_{e,i,j})\cdot x_{e}^{(t)}$.

\begin{figure*}[h!]
\vskip -0.05in \centering \includegraphics[width=1\linewidth]{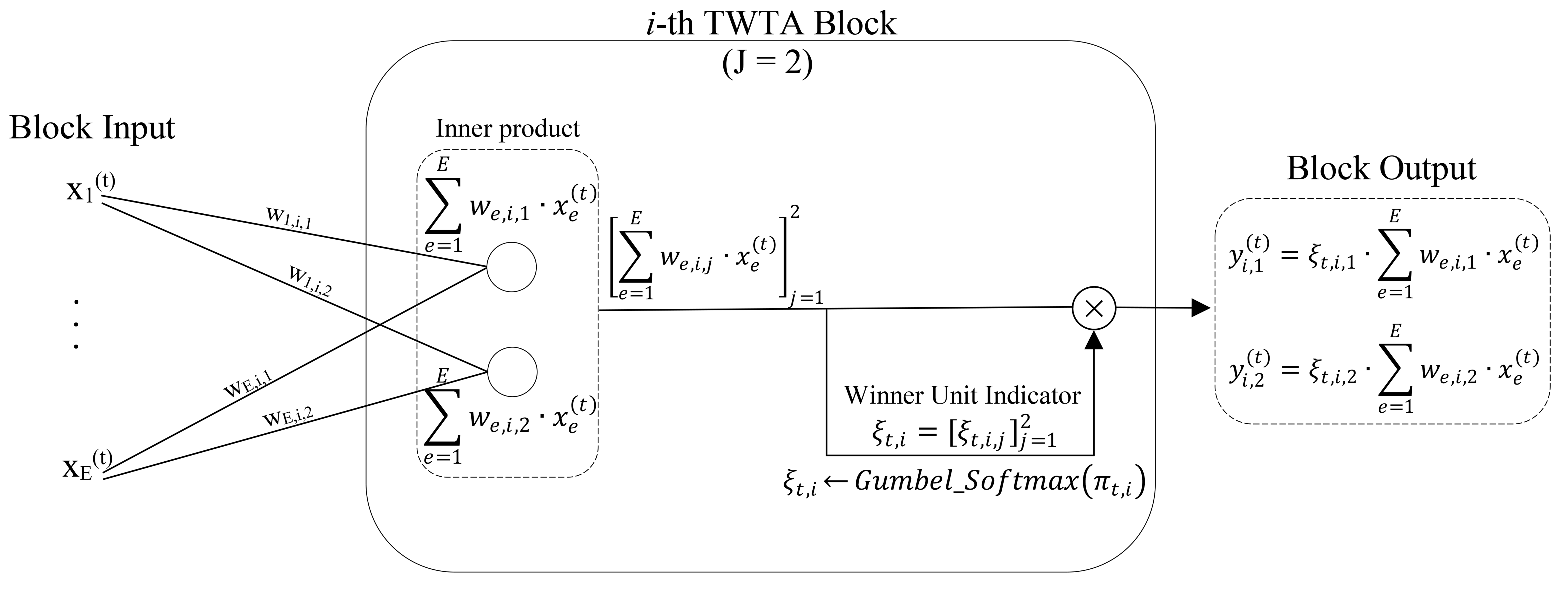}
\caption{A detailed graphical illustration of the $i$-th block of a proposed
TWTA layer (Section \ref{model_form}); for demonstration purposes,
we choose $J=2$ competing units per block. Inputs $\bm{x}^{(t)}=\{x_{1}^{(t)},\dots,x_{E}^{(t)}\}$
are presented to each unit in the $i$-th block, when training on
task $t$. Due to the TWTA mechanism, during forward passes through
the network, only one competing unit propagates its output to the
next layer; the rest are zeroed-out.}
\label{fig:arch_dense} \vskip -0.05in 
\end{figure*}

Fig. \ref{fig:arch_dense} illustrates the operation of the proposed
architecture when dealing with task $t$. As we observe, for each
task only one unit (the ``winner'') in a TWTA block will present
its output to the next layer during forward passes through the network;
the rest are zeroed out. During backprop (training), the strength
of the updating signal is regulated from the relaxed (continuous)
outcome of the competition process; this is encoded into a (differentiable)
sample from the postulated Gumbel-Softmax.

\begin{figure*}[h!]
\vskip -0.05in \centering \includegraphics[scale=0.1]{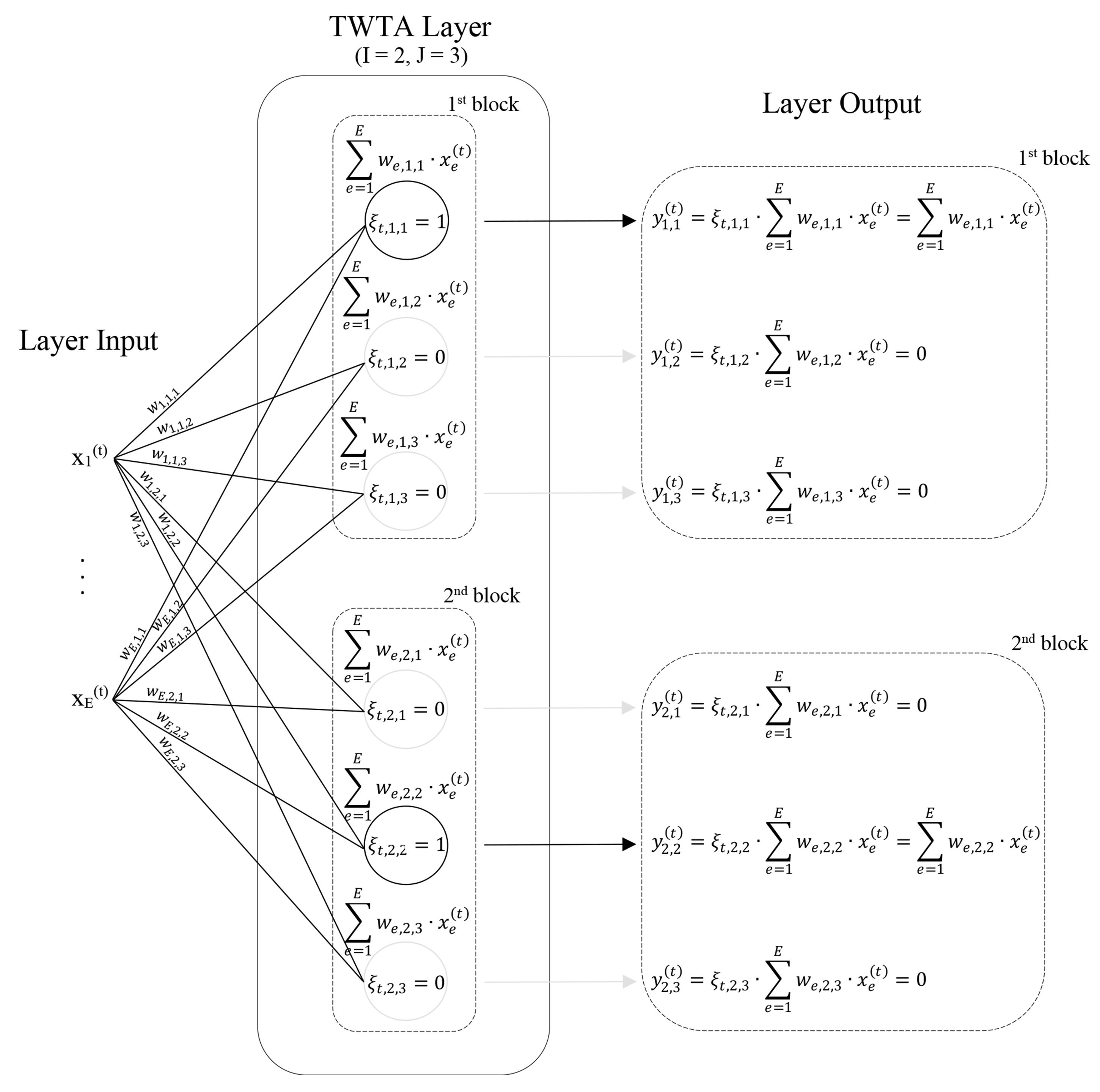}
\caption{A detailed graphical illustration of a TWTA layer (Section \ref{model_form});
for demonstration purposes, we choose $I=2$ blocks with $J=3$ competing
units per block. Inputs $\bm{x}^{(t)}=\{x_{1}^{(t)},\dots,x_{E}^{(t)}\}$
are presented to each unit in all blocks, when training on task $t$. }
\label{fig:arch_dense-layer} \vskip -0.05in
\end{figure*}

Fig. \ref{fig:arch_dense-layer} illustrates a full TWTA layer, comprising
$I$ blocks with $J$ competing units each. We introduce the hidden
winner indicator vector $\bm{\xi}_{t,i}=[\xi_{t,i,j}]_{j=1}^{J}$
of the $i$-th block pertaining to the $t$-th task. It holds $\xi_{t,i,j}=1$
if the $j$-th unit in the $i$-th block has specialized in the $t$-th
task (winning unit), $\xi_{t,i,j}=0$ otherwise. We also denote $\bm{\xi}_{t}\in\{0,1\}^{I\cdot J}$
the vector that holds all the $\bm{\xi}_{t,i}\in\{0,1\}^{J}$ subvectors.
On this basis, the output of the layer, $\bm{y}^{(t)}\in\mathbb{R}^{I\cdot J}$,
in our approach is composed of $I$ sparse subvectors $\bm{y}_{i}^{(t)}\in\mathbb{R}^{J}$.
Succinctly, we can write $\bm{y}_{i}^{(t)}=[{y}_{i,j}^{(t)}]_{j=1}^{J}$,
where: 
\begin{equation}
{y}_{i,j}^{(t)}=\xi_{t,i,j}\sum_{e=1}^{E}(w_{e,i,j})\cdot x_{e}^{(t)}\in\mathbb{R}\label{out_y}
\end{equation}

We postulate that the hidden winner indicator variables are drawn
from a Categorical posterior distribution that yields: 
\begin{equation}
p(\bm{\xi}_{t,i})=\mathrm{Categorical}(\bm{\xi}_{t,i}|\bm{\pi}_{t,i})\label{eq:categor}
\end{equation}

The hyperparameters $\bm{\pi}_{t,i}$ are optimized during model training,
as we explain next. The network learns the global weight matrix $\bm{W}$,
that is not specific to a task, but evolves over time. During training,
by learning different winning unit distributions, $p(\bm{\xi}_{t,i})$,
for each task, we appropriately mask-out large parts of the network,
dampen the training signal strength for these parts, and mainly direct
the training signal to update only a fraction of $\bm{W}$ that pertains
to the remainder of the network. This results in a very slim weight
updating scheme during backpropagation, which updates only a small
winning subnetwork; thus, we yield important computational savings
of immense significance to edge devices.

In Fig. \ref{fig:arch_dense}, we depict the operation principles
of our proposed network. As we show therein, the winning information
is encoded into the trained posteriors $p(\bm{\xi}_{t,i})$, which
are used to regulate weight training, as we explain in Section \ref{training}.
This is radically different from \cite{Chen_Zhang}, as we do not
search for optimal winning tickets during CIL via repetitive pruning
and retraining for each arriving task. This is also radically different
from \cite{Kang_Yoon}, where a random uniform mask is drawn for regulating
which weights will be updated, and another mask is optimized to select
the subnetwork specializing to the task at hand. Instead, we perform
a single backward pass to update the winning unit distributions, $p(\bm{\xi}_{t,i})$,
and the weight matrix, $\bm{W}$; importantly, the updates of the
former (winning unit posteriors) regulate the updates of the latter
(weight matrix).

\textbf{Inference}. As we depict in Fig. \ref{fig:arch_dense}, during
inference for a given task $t$, we retain the unit with maximum hidden
winner indicator variable posterior, $\pi_{t,i,j}$, in each TWTA
block $i$, and prune-out the weights pertaining to the remainder
of the network. The feedforward pass is performed, by definition,
by computing the task-wise discrete masks: 
\begin{equation}
\begin{aligned}\tilde{\bm{\xi}}_{t,i}= & \mathrm{onehot}\left(\arg\max\limits _{j}\pi_{t,i,j}\right)\in\mathbb{R}^{J}\\
\textrm{Thus: } & \mathrm{winner}_{t,i}\triangleq\mathrm{arg}\,\underset{j}{\mathrm{max}}\,\pi_{t,i,j}
\end{aligned}
\label{eq:discrete_mask}
\end{equation}
Apparently, this way the proportion of retained weights for task $t$
is only equal to the $\frac{1}{J}*100\%$ of the number of weights
the network is initialized with.

\subsection{A Convolutional Variant}

\label{conv_var} 
\begin{figure*}[h!]
\vskip -0.075in \centering \includegraphics[width=1\linewidth]{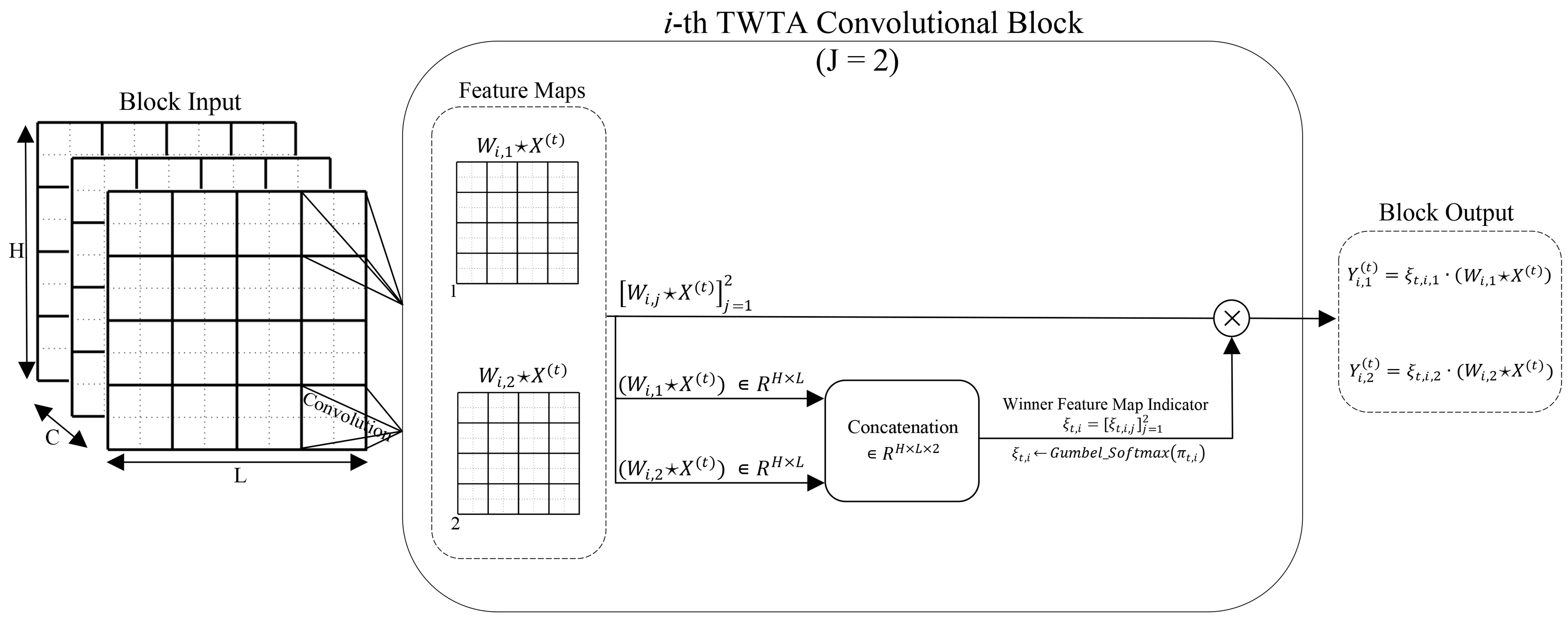}
\caption{The convolutional TWTA variant (Section \ref{conv_var}); for demonstration
purposes, we choose $J=2$ competing feature maps per kernel. Due
to the TWTA mechanism, during forward passes through the network,
only one competing feature map propagates its output to the next layer;
the rest are zeroed-out. }
\label{fig:arch_conv} \vskip -0.1in 
\end{figure*}

Further, to accommodate architectures comprising
convolutional operations, we consider a variant of the TWTA layer,
inspired from \cite{Panousis_Chatzis}. In the remainder of this work,
this will be referred to as the Conv-TWTA layer, while the original
TWTA layer will be referred to as the dense variant. The graphical
illustration of Conv-TWTA is provided in Fig. \ref{fig:arch_conv}.

Specifically, let us assume an input tensor $\bm{X}^{(t)}\in\mathbb{R}^{H\times L\times C}$
of a layer, where $H,L,C$ are the height, length and channels of
the input. We define a set of kernels, each with weights $\bm{W}_{i}\in\mathbb{R}^{h\times l\times C\times J}$,
where $h,l,C,J$ are the kernel height, length, channels and competing
feature maps, and $i=1,\dots,I$.

Here, analogously to the grouping of linear units in a dense TWTA
layer of Section \ref{model_form}, local competition is performed
among feature maps in a kernel. Thus, each kernel is treated as a
TWTA block, feature maps in a kernel compete among them, and multiple
kernels of competing feature maps constitute a Conv-TWTA layer.

This way, the output $\bm{Y}^{(t)}\in\mathbb{R}^{H\times L\times(I\cdot J)}$
of a layer under the proposed convolutional variant is obtained via
concatenation along the last dimension of the subtensors $\bm{Y}_{i}^{(t)}$:
\begin{equation}
\bigskip\bm{Y}_{i}^{(t)}=\bm{\xi}_{t,i}\cdot(\bm{W}_{i}\star\bm{X}^{(t)})\in\mathbb{R}^{H\times L\times J}
\label{eq:conv}
\end{equation}
where ``$\star$'' denotes the convolution operation.

Here, the winner indicator variables $\bm{\xi}_{t,i}$ are drawn again
from the distribution of Eq. (\ref{eq:categor}); they now govern
competition among feature maps of a kernel.

\textbf{Inference}. At inference time, we employ a similar winner-based
weight pruning strategy as for dense TWTA layers; for each task, the
weights associated with one feature map (the winner) are retained
while the rest are zeroed-out. Specifically, the winning feature map
in a kernel $i$ for task $t$ is selected through $\mathrm{arg\,max}$
over the hidden winner indicator variable posteriors, $\pi_{t,i,j}\,\forall j$,
similar to Eq. (\ref{eq:discrete_mask}). (see also Fig. \ref{fig:arch_conv}).
This massively sparsifies weights, rendering the network amenable
to edge devices.

\subsection{Training}

\label{training} For each task $t$, our approach consists in executing
\emph{a single} full training cycle. The performed training cycle
targets both the network weights, $\bm{W}$, and the posterior hyperparameters
$\bm{\pi}_{t,i}$ of the winner indicator hidden variables pertaining
to the task, $p(\bm{\xi}_{t,i})\;\forall i$.

The vectors $\bm{\pi}_{t,i}$ are initialized at random, while the
network weights, $\bm{W}$, ``continue'' from the estimator obtained
after training on task $t-1$. We denote as $\bm{W}^{(t)}$ the updated
weights estimator obtained through the training cycle on the $t$-th
task.

To perform training, we resort to minimization of a simple categorical
cross-entropy criterion. Let us consider the $u$-th training iteration
on task $t$, with data batch $D_{u}^{(t)}=(X_{u}^{(t)},Y_{u}^{(t)})$.
The training criterion is the categorical cross-entropy $\mathrm{CE}(Y_{u}^{(t)},f(X_{u}^{(t)};\bm{W}^{(t)},\hat{\boldsymbol{\xi}_{t}}))$
between the data labels $Y_{u}^{(t)}$ and the class probabilities
$f(X_{u}^{(t)};\bm{W}^{(t)},\hat{\boldsymbol{\xi}_{t}})$ generated
from the penultimate Softmax layer of the network. In this definition,
$\hat{\boldsymbol{\xi}}_{t}=[\hat{\boldsymbol{\xi}}_{t,i}]_{i=1}^{I}$
is a vector concatenation of single Monte-Carlo (MC) samples drawn
from the Categorical posteriors $p(\boldsymbol{\xi}_{t,i})$.

To ensure low-variance gradients with only one drawn MC sample, we
reparameterize these samples by resorting to the Gumbel-Softmax relaxation
\cite{Maddison}. The Gumbel-Softmax relaxation yields sampled instances
$\hat{\bm{\xi}}_{t,i}$ under the following expression: 
\begin{equation}
\begin{split} & \hat{\bm{\xi}}_{t,i}=\mathrm{Softmax}(([\log{\pi}_{t,i,j}+g_{t,i,j}]_{j=1}^{J})/\tau)\in\mathbb{R}^{J},\;\forall i\\
 & \mathrm{where:}\;g_{t,i,j}=-\log(\,-\log U_{t,i,j}),\;U_{t,i,j}\sim\mathrm{Uniform}(0,1)
\end{split}
\label{eq:GS}
\end{equation}
and $\tau\in(0,\infty)$ is a temperature factor that controls how
closely the Categorical distribution $p(\bm{\xi}_{t,i})$ is approximated
by the continuous relaxation. This is similar to \cite{Panousis_Chatzis}.

\section{Related Work}

\label{related_work}

Recent works in \cite{Chen_Zhang,Kang,Kang_Yoon} have pursued to
build computationally efficient continual learners by drawing inspiration
from LTH \cite{Frankle}. These works compose sparse subnetworks that
achieve comparable or/and even higher predictive performance than
their initial counterparts. However, our work is substantially different
from the existing state-of-the-art, as it specifically accounts for
the constraints imposed in the context of execution on an edge device:
\\
 (i) Contrary to \cite{Chen_Zhang}, we do not employ iterative pruning,
which repeats multiple full cycles of network training and pruning;
this would be completely intractable on an edge device. Instead, we
perform a single training cycle, at the end of which we retain a (task-specific)
subnetwork to perform inference for the task. \\
 (ii) \cite{Kang} and \cite{Kang_Yoon} select a subnetwork that
will be used for the task at hand on the grounds of an optimization
criterion for binary masks imposed over the network weights. Once
this subnetwork has been selected, they train randomly selected subsets
of the weights of the whole network, to account for the case of a
suboptimal subnetwork selection. Similar to \cite{Chen_Zhang}, this
is completely intractable on an edge device.

On the contrary, our method attempts to encourage different units
in a competing block to specialize to different tasks. Training is
performed concurrently for the winner unit indicator hidden variables,
the posteriors of which regulate weight updates, as well as the network
weights themselves. Thus, network pruning comes at the end of weight
updating and not beforehand. We posit that this regulated updating
scheme, which does not entail a priori hard pruning decisions, facilitates
generalization without harming catastrophic forgetting. 

On the other hand, \cite{Wang:2022aa} are the first to directly attack
the problem of CL on an edge device. To this end, they suggest an
weight importance metric and a related weight gradient importance
metric, which attempt to retain (1) weights of larger magnitude for
output stability, (2) weights important for the current task for learning
capacity, and (3) weights important for past data to mitigate catastrophic
forgetting. However, the use of these metrics requires the application
of intra-task and inter-task adjustment processes, which require extensive
heuristic tuning. In addition, the use of two different but related
metrics for the weights and their gradients is due to the heuristic
thresholds this method requires, which cannot be homogeneous. This
is a drawback that makes the method hard to use off-the-shelf in a
given scenario. Finally, for the method to 

\section{Experiments}

To demonstrate the effectiveness of our method for edge devices, we
conducted experiments using a simulated edge computing environment.
We benchmarked our sparse model against traditional dense models on
various datasets, observing performance in terms of computational
efficiency and memory usage.

\label{experiments} We evaluate on CIFAR-100 \cite{Krizhevsky},
Tiny-ImageNet \cite{Stanford}, PMNIST \cite{LeCun_pmnist} and Omniglot
Rotation \cite{Lake}. Also, we evaluate on the 5-Datasets \cite{Saha}
benchmark, in order to examine how our method performs in case that
cross-task generalization concerns different datasets. We randomly
divide the classes of each dataset into a fixed number of tasks with
a limited number of classes per task. Specifically, in each training
iteration, we construct $N$-way few-shot tasks by randomly picking
$N$ classes and sampling few training samples for each class. In
Supplementary Section \ref{appendix_exp_details}, we specify further
experimental details for our datasets.

We adopt the original ResNet18 network \cite{He} for Tiny-ImageNet,
PMNIST and 5-Datasets; we use a 5-layer AlexNet similar to \cite{Saha}
for the experiments on CIFAR-100, and LeNet \cite{LeCun_pmnist} for
Omniglot Rotation. In the case of our approach, we modify those baselines
by replacing each ReLU layer with a layer of (dense) TWTA blocks,
and each convolutional layer with a layer of Conv-TWTA blocks. See
more details in the Supplementary Section \ref{appendix_arch}.

For both the network weights, $\bm{W}$, and the log hyperparameters,
$\log\bm{\pi}_{t,i}$, we employ Glorot Normal initialization \cite{glorot}.
At the first training iteration of a new task, we initialize the Gumbel-Softmax
relaxation temperature $\tau$ to 0.67; as the training proceeds,
we linearly anneal its value to 0.01. We use SGD optimizer \cite{Robbins}
with a learning rate linearly annealed to 0, and initial value of
0.1. We run 100 training epochs per task, with batch size of 40.

\subsection{Experimental results}

\label{exp_results}

\begin{table*}[h!]
\centering 
\caption{Comparisons on CIFAR-100, Tiny-ImageNet, PMNIST, Omniglot Rotation
and 5-Datasets. We set $J=32$; thus, the proportion of retained weights
for each task, after training, is equal to the $(\frac{1}{J}*100=3.125)\%$
of the initial network. Also, we show the number of retained weights
after training (in millions), for our method and the alternative approaches
for reducing model size.}
\label{results1} 
\resizebox{\linewidth}{3.5cm}{ %
\begin{tabular}{lccccc}
\toprule 
\textbf{Algorithm}  & \textbf{CIFAR-100}  & \textbf{Tiny-ImageNet}  & \textbf{PMNIST}  & \textbf{Omniglot Rotation}  & \textbf{5-Datasets} \tabularnewline
\midrule 
GEM \cite{Lopez}  & 59.24 & 39.12 & -  & -  & - \tabularnewline
iCaRL \cite{Rebuffi}  & 42.45 & 43.97 & 55.82 & 44.60 & 48.01\tabularnewline
ER \cite{Chaudhry}  & 59.12 & 37.65 & -  & -  & - \tabularnewline
IL2M \cite{Belouadah}  & 53.24 & 47.13 & 60.12 & 51.31 & 55.93\tabularnewline
La-MAML \cite{Gupta}  & 60.02 & 55.45 & 80.82 & 61.73 & 75.92\tabularnewline
FS-DGPM \cite{Deng}  & 63.81  & 59.74 & 80.92 & 62.83 & 76.10\tabularnewline
GPM \cite{Saha}  & 62.40  & 56.28 & 83.51 & 74.63 & 80.75\tabularnewline
\midrule 
SoftNet (80$\%$, 4.5M params)  & 48.52  & 54.02 & 64.02 & 55.82 & 57.60\tabularnewline
SoftNet (10$\%$, 0.69M params)  & 43.61  & 47.30 & 57.93 & 46.83 & 52.11\tabularnewline
LLT (100$\%$, 11M params)  & 61.46  & 58.45 & 80.38 & 70.19 & 74.61\tabularnewline
LLT (6.87$\%$, 0.77M params)  & 62.69  & 59.03 & 80.91 & 68.46 & 75.13\tabularnewline
WSN (50$\%$, 4.2M params)  & 64.41 & 57.83 & 84.69 & 73.84 & 82.13\tabularnewline
WSN (8$\%$, 0.68M params)  & 63.24 & 57.11 & 83.03 & 72.91 & 79.61\tabularnewline
\midrule 
\textbf{TWTA-CIL (3.125$\%$, 0.0975M params)}  & \textbf{66.53} & \textbf{61.93} & \textbf{85.92} & \textbf{76.48} & \textbf{83.77}\tabularnewline
\midrule 
SparCL (5\%, 0.54M params) & 59.55 & 52.16 & 76.82 & 66.53 & 72.84\tabularnewline
\bottomrule
\end{tabular}} 
\end{table*}

\begin{table*}[h!]
\vskip -0.05in \centering 
\caption{Average training wall-clock time (in secs), c.f. Table \ref{results1}.}
\label{results1_time} 
\resizebox{\linewidth}{1cm}{ %
\begin{tabular}{lccccc}
\toprule 
\textbf{Algorithm}  & \textbf{CIFAR-100}  & \textbf{Tiny-ImageNet}  & \textbf{PMNIST}  & \textbf{Omniglot Rotation}  & \textbf{5-Datasets} \tabularnewline
\midrule 
\textbf{TWTA-CIL (3.125$\%$, 0.0975M params)}  & $3.7\times10^{15}$ & $3\times10^{15}$ & $5.4\times10^{15}$ & $6.1\times10^{15}$ & $8.5\times10^{15}$\tabularnewline
\midrule 
SparCL (5\%, 0.54M params) & $1.2\times10^{16}$ & $1.\times10^{16}$ & $3.1\times10^{16}$ & $4.4\times10^{16}$ & $9.3\times10^{16}$\tabularnewline
\bottomrule
\end{tabular}} \vskip -0.05in 
\end{table*}

\begin{table*}[h!]
\centering 
\caption{BTI over the considered algorithms and datasets of Table \ref{results1};
the lower the better.}
\label{BTI_table} 
\begin{tabular}{lccccc}
\toprule 
\textbf{Algorithm}  & \textbf{CIFAR-100}  & \textbf{Tiny-ImageNet}  & \textbf{PMNIST}  & \textbf{Omniglot Rotation}  & \textbf{5-Datasets} \tabularnewline
\midrule 
iCaRL  & 13.41  & 6.45  & 8.51  & 18.41  & 23.56 \tabularnewline
IL2M  & 20.41  & 7.61  & 9.03  & 14.60  & 19.14 \tabularnewline
La-MAML  & 7.84  & 13.84  & 10.51  & 17.04  & 15.13 \tabularnewline
FS-DGPM  & 9.14  & 12.25  & 8.85  & \textbf{13.64}  & 19.51 \tabularnewline
GPM  & 12.44  & 8.03  & 11.94  & 16.39  & 17.11 \tabularnewline
\midrule 
SoftNet (80$\%$)  & 13.80  & 9.62  & 10.38  & 18.12  & 18.04 \tabularnewline
SoftNet (10$\%$)  & 12.09  & 8.33  & 9.76  & 16.30  & 18.68 \tabularnewline
LLT (100$\%$)  & 15.02  & 7.05  & 9.54  & 15.31  & 14.80 \tabularnewline
LLT (6.87$\%$)  & 14.61  & 3.51  & 11.84  & 17.12  & 17.46 \tabularnewline
WSN (50$\%$)  & 11.14  & 4.81  & 10.51  & 14.20  & 20.41 \tabularnewline
WSN (8$\%$)  & 10.58  & 8.78  & 9.32  & 15.34  & 18.92 \tabularnewline
\midrule 
\textbf{TWTA-CIL (12.50$\%$)}  & \textbf{6.14}  & \textbf{2.50}  & \textbf{8.04}  & \textbf{13.64}  & \textbf{13.51} \tabularnewline
\bottomrule
\end{tabular} \vskip -0.1in 
\end{table*}

\begin{table*}[h!]
\caption{Effect of block size $J$; Tiny-ImageNet and CIFAR-100 datasets. The
higher the block size $J$ the lower the fraction of the trained network
retained at inference time.}
\label{ablation_effect_j} 
\centering {\small{}{}
}%
\begin{tabular}{lcccccc}
\toprule 
 & \multicolumn{3}{c}{\textbf{\small{}{}Tiny-ImageNet}} & \multicolumn{3}{c}{\textbf{\small{}{}CIFAR-100}}\tabularnewline
\midrule 
\textbf{\small{}{}Algorithm}{\small{}{} }  & {\small{}{}Time }  & {\small{}{}Accuracy }  & {\small{}{}J }  & {\small{}{}Time }  & {\small{}{}Accuracy }  & {\small{}{}J }\tabularnewline
\midrule 
{\small{}{}TWTA-CIL (50$\%$) }  & {\small{}{}2634.02 }  & {\small{}{}61.32} & {\small{}{}2 }  & {\small{}{}1493.79 }  & {\small{}{}65.73}  & {\small{}{}2 }\tabularnewline
{\small{}{}TWTA-CIL (25$\%$) }  & {\small{}{}2293.81 }  & {\small{}{}61.04} & {\small{}{}4 }  & {\small{}{}1301.20 }  & {\small{}{}65.45} & {\small{}{}4 }\tabularnewline
\textbf{\small{}{}TWTA-CIL (12.50$\%$)}{\small{}{} }  & {\small{}{}1914.63 }  & \textbf{\small{}{}61.93} & {\small{}{}8 }  & {\small{}{}1039.73 }  & \textbf{\small{}{}66.53} & {\small{}{}8 }\tabularnewline
{\small{}{}TWTA-CIL (6.25$\%$) }  & {\small{}{}1556.09 }  & {\small{}{}61.45} & {\small{}{}16 }  & {\small{}{}801.46 }  & {\small{}{}65.89} & {\small{}{}16 }\tabularnewline
{\small{}{}TWTA-CIL (3.125$\%$) }  & {\small{}{}1410.64 }  & {\small{}{}60.86} & {\small{}{}32 }  & {\small{}{}785.93 }  & {\small{}{}65.40} & {\small{}{}32 }\tabularnewline
\bottomrule
\end{tabular}{\small{}{}
} 
\end{table*}

In Table \ref{results1}, we show how TWTA-CIL performs in various
benchmarks compared to popular alternative methods. We emphasize that
the performance of SoftNet and WSN is provided for the configuration
reported in the literature that yields the best accuracy, as well
as for the reported configuration that corresponds to the proportion
of retained weights closest to our method. Turning to LLT, we report
how the method performs with no pruning and with pruning ratio closest
to our method.

As we observe, our method outperforms the existing state-of-the-art
in every considered benchmark. For instance, WSN performs worse than
TWTA-CIL (3.125\%), irrespectively of whether WSN retains a greater
or a lower proportion of the initial network. Thus, our approach successfully
discovers sparse subnetworks (\textit{winning tickets}) that are powerful
enough to retain previous knowledge, while generalizing well to new
unseen tasks. Crucially, our method outperforms all the alternatives,
including the related, edge device-oriented SparCL approach, in terms
of both obtained accuracy and number of retained parameters at inference
time (thus, memory footprint). 

Finally, it is interesting to examine how the winning ticket vectors
differentiate across tasks. To this end, we compute the overlap among
the $\tilde{\bm{\xi}}_{t}=[\tilde{\bm{\xi}}_{t,i}]_{i}$ vectors,
defined in Eq. (\ref{eq:discrete_mask}), for all consecutive pairs
of tasks, $(t-1,t)$, and compute average percentages. We observe
that average overlap percentages range from $6.38\%$ to $10.67\%$
across the considered datasets; this implies clear differentiation.

\subsubsection{Computational times for training CIL methods}

\label{section_times}

In Table \ref{results1_time}, we report the training FLOPs for our
method and its direct competitor, that is SparCL \cite{Wang:2022aa}.
It is apparent that our method yields much improved training algorithm
computational costs.

\subsubsection{Reduction of forgetting tendencies}

\label{ablation}

To examine deeper the obtained improvement in forgetting tendencies,
we report the \textit{backward-transfer and interference} (BTI) values
of the considered methods in Table \ref{BTI_table}. BTI measures
the average change in the accuracy of each task from when it was learnt
to the end of the training, that is training on the last task; thus,
it is immensely relevant to this empirical analysis. A smaller value
of BTI implies lesser forgetting as the network gets trained on additional
tasks. As Table \ref{BTI_table} shows, our approach forgets less
than the baselines on all benchmarks.

\subsection{Effect of block size $J$}

\label{section_effect_j} Finally, we re-evaluate TWTA-CIL with various
block size values $J$ (and correspondingly varying number of layer
blocks, $I$). In all cases, we ensure that the total number of feature
maps, for a convolutional layer, or units, for a dense layer, which
equals $I*J$, remains the same as in the original architecture of
Section \ref{exp_results}. This is important, as it does not change
the total number of trainable parameters, but only the organization
into blocks under the local winner-takes-all rationale. Different
selections of $J$ result in different percentages of remaining network
weights at inference time, as we can see in Table \ref{ablation_effect_j}
(datasets Tiny-ImageNet and CIFAR-100). As we observe, the ``TWTA-CIL
(12.50\%)'' alternative, with $J=8$, is the most accurate configuration
of TWTA-CIL. However, the most efficient version, perfectly fit for
application to edge devices, {\small{}that is TWTA-CIL (3.125$\%$),
yields only a negligible accuracy} drop in all the considered benchmarks.{\small{} }{\small\par}

\section{Conclusion}

This paper presented a sparsity-promoting method tailored for continual
learning on edge devices. By incorporating stochastic competition,
we achieved an approach that is both efficient and effective, suitable
for the limited capabilities of edge computing. Specifically, the
results clearly demonstrated that our sparsity-promoting method significantly
outperforms traditional models on edge devices. We observed a significant
reduction in memory usage and an increase in computational speed,
confirming the suitability of our approach for deployment in resource-constrained
environments. Future research may explore further optimizations to
enhance the adaptability of this method across more diverse edge computing
scenarios.

\bibliographystyle{icml2024}
\bibliography{iclr2024_conference}

\appendix

\begin{table*}[h!]
\centering 
\caption{Modified ResNet18 architecture parameters.}
\label{arch_j} %
\begin{tabular}{lccccc}
\toprule 
\textbf{Layer Type} & $(J=2)$ & $(J=4)$ & $(J=8)$ & $(J=16)$ & $(J=32)$\tabularnewline
\midrule 
TWTA-Conv & 8 & 4 & 2 & 1 & 1\tabularnewline
\midrule 
4x TWTA-Conv & 8 & 4 & 2 & 1 & 1\tabularnewline
4x TWTA-Conv & 8 & 4 & 2 & 1 & 1\tabularnewline
4x TWTA-Conv & 16 & 8 & 4 & 2 & 1\tabularnewline
4x TWTA-Conv & 16 & 8 & 4 & 2 & 1\tabularnewline
\midrule 
TWTA-Dense & 16 & 8 & 4 & 2 & 1\tabularnewline
\bottomrule
\end{tabular}

\end{table*}

\begin{table*}[h!]
\centering 
\caption{Modified AlexNet architecture parameters.}
\label{alex_arch} %
\begin{tabular}{lccccc}
\toprule 
\textbf{Layer Type} & $(J=2)$ & $(J=4)$ & $(J=8)$ & $(J=16)$ & $(J=32)$\tabularnewline
\midrule 
TWTA-Conv & 8 & 4 & 2 & 1 & 1\tabularnewline
\midrule 
TWTA-Conv & 8 & 4 & 2 & 1 & 1\tabularnewline
TWTA-Conv & 8 & 4 & 2 & 1 & 1\tabularnewline
TWTA-Conv & 16 & 8 & 4 & 2 & 1\tabularnewline
TWTA-Dense & 64 & 32 & 16 & 8 & 4\tabularnewline
\midrule 
TWTA-Dense & 64 & 32 & 16 & 8 & 4\tabularnewline
\bottomrule
\end{tabular}
\end{table*}

\section{More details on the used datasets}

\label{appendix_exp_details}

\textbf{Datasets and Task Splittings}\;\: 5-Datasets is a mixture
of 5 different vision datasets: CIFAR-10, MNIST \cite{LeCun_pmnist},
SVHN \cite{Netzer}, FashionMNIST \cite{Xiao} and notMNIST \cite{Bui}.
Each dataset consists of 10 classes, and classification on each dataset
is treated as a single task. PMNIST is a variant of MNIST, where each
task is generated by shuffling the input image pixels by a fixed permutation.
In the case of Omniglot Rotation, we preprocess the raw images of
Omniglot dataset by generating rotated versions of ($90^{\circ},180^{\circ},270^{\circ}$)
as in \cite{Kang}. For 5-Datasets, similar to \cite{Kang}, we pad
0 values to raw images of MNIST and FashionMNIST, convert them to
RGB format to have a dimension of 3$*$32$*$32, and finally normalize
the raw image data. All datasets used in Section \ref{experiments}
were randomly split into training and testings sets with ratio of
9:1. The number of stored images in the memory buffer - per class
- is 5 for Tiny-ImageNet, and 10 for CIFAR-100, PMNIST, Omniglot Rotation
and 5-Datasets.

We randomly divide the 100 classes of CIFAR-100 into 10 tasks with
10 classes per task; the 200 classes of Tiny-ImageNet into 40 tasks
with 5 classes per task; the 200 classes of PMNIST into 20 tasks with
10 classes per task; and in the case of Omniglot Rotation, we divide
the available 1200 classes into 100 tasks with 12 classes per task.
The $N$-way few-shot settings for the constructed tasks in each training
iteration are: $10$-way $10$-shot for CIFAR-100, PMNIST and 5-Datasets,
$5$-way $5$-shot for Tiny-ImageNet, and $12$-way $10$-shot for
Omniglot Rotation.

\textbf{Unlabeled Dataset}\;\: The external unlabeled data are retrieved
from 80 Million Tiny Image dataset \cite{Torralba} for CIFAR-100,
PMNIST, Omniglot Rotation and 5-Datasets, and from ImageNet dataset
\cite{Krizhevsky} for Tiny-ImageNet. We used a fixed buffer size
of 128 for querying the same number of unlabeled images per class
of learned tasks at each training iteration, based on the feature
similarity that is defined by $l_{2}$ norm distance.

\section{Modified Network Architecture details}

\label{appendix_arch}

\subsection{ResNet18}

The original ResNet18 comprises an initial convolutional layer with
64 3x3 kernels, 4 blocks of 4 convolutional layers each, with 64 3x3
kernels on the layers of the first block, 128 3x3 kernels for the
second, 256 3x3 kernels for the third and 512 3x3 kernels for the
fourth. These layers are followed by a dense layer of 512 units, a
pooling and a final Softmax layer. In our modified ResNet18 architecture,
we consider kernel size = 3 and padding = 1; in Table \ref{arch_j},
we show the number of used kernels / blocks and competing feature
maps / units, $J$, in each modified layer.

\subsection{AlexNet}

The 5-layer AlexNet architecture comprises 3 convolutional layers
of 64, 128, and 256 filters with 4x4, 3x3, and 2x2 kernel sizes, respectively.
These layers are followed by two dense layers of 2048 units, with
rectified linear units as activations, and 2x2 max-pooling after the
convolutional layers. The final layer is a fully-connected layer with
a Softmax output. In our modified AlexNet architecture, we replace
each dense ReLU layer with a layer of (dense) TWTA blocks, and each
convolutional layer with a layer of Conv-TWTA blocks; in Table \ref{alex_arch},
we show the number of used kernels / blocks and competing feature
maps / units, $J$, in each modified layer.

\subsection{LeNet}

The LeNet architecture comprises 2 convolutional layers of 20, and
50 feature maps, followed by one feedforward fully connected layer
of 500 units, and a final Softmax layer. In our modified LeNet architecture,
we replace each of the 2 convolutional layers with one layer of Conv-TWTA
blocks; the former retains 2 kernels / blocks of 8 competing feature
maps, and the latter 6 kernels / blocks of 8 competing feature maps.
The fully connected layer is replaced with a dense TWTA-layer, consisting
of 50 blocks of 8 competing units.

\end{document}